\documentclass[11pt]{article}

\usepackage{geometry}
\usepackage{setspace}
\usepackage{caption} 
\usepackage{graphicx}
\usepackage{mathptmx}
\usepackage[T1]{fontenc}

\newif\ifbis
\bisfalse 

\ifbis
\usepackage[natbibapa]{apacite}
\newenvironment{biseabstract}{%
\begin{quote} \bf}
{\end{quote}}

\newenvironment{bisekeywords}{%
\begin{quote} \it \textbf{Keywords -}}
{\end{quote}}

\newcommand{\ci}{\citep}
\else 

\usepackage{url}
\newcommand{\ci}{\cite}
\fi

\ifbis
\title{Foundation models} 
\else
\title{Foundation models in brief: A historical, socio-technical focus\footnote{This is a pre-print of a submitted manuscript.}} 
\fi

\author{Johannes Schneider\\
\\
\normalsize{Institute of Information Systems, University of Liechtenstein}\\
\normalsize{Fuerst Franz Josef Str., 9490 Vaduz, Liechtenstein}\\
\normalsize{E-mail: johannes.schneider@uni.li}
}

\date{}

\begin{document} 
\maketitle 

\ifbis
\begin{biseabstract}
\else
\begin{abstract}
\noindent
\fi
Foundation models can be disruptive for future AI development by scaling up deep learning in terms of model size and training data's breadth and size. These models achieve state-of-the-art performance (often through further adaptation) on a variety of tasks in domains such as natural language processing and computer vision. Foundational models exhibit a novel \emph{emergent behavior}: \emph{In-context learning} enables users to provide a query and a few examples from which a model derives an answer without being trained on such queries. Additionally, \emph{homogenization} of models might replace a myriad of task-specific models with fewer very large models controlled by few corporations leading to a shift in power and control over AI. This paper provides a short introduction to foundation models. It contributes by crafting a crisp distinction between foundation models and prior deep learning models, providing a history of machine learning leading to foundation models, elaborating more on socio-technical aspects, i.e.,  organizational issues and end-user interaction, and a discussion of future research.  

\ifbis
\end{biseabstract}
\begin{bisekeywords}
Foundation Models, Artificial Intelligence, Emergent Behavior, Homogenization, Transformers 
\end{bisekeywords}
\baselineskip24pt
\else
\end{abstract}
\fi

\section*{Introduction}
Current foundation models\ci{bom21} are large-scale artificial intelligence models, i.e., deep learning models, trained on a large amount of broad, typically unlabeled data. They cover well-known models popularized in the press, such as GPT-3 \ci{bro20}, Dall-E 2 \ci{ram22}, Florence \ci{yua21} and, widely adapted, early models, such as BERT \ci{dev18}.  They arguably constitute the next milestone in the evolution of the main branch of AI, although they still face many diverse challenges and come with substantial risks spanning many socio-technical concerns. In turn, interdisciplinary research is needed. For illustration, Stanford university alone has established the Center for Research on Foundation Models covering more than ten departments and more than 100 researchers contributing to this topic\ci{bom21}.

\begin{figure}[htb]
    \centering
    \includegraphics[width=0.6\textwidth]{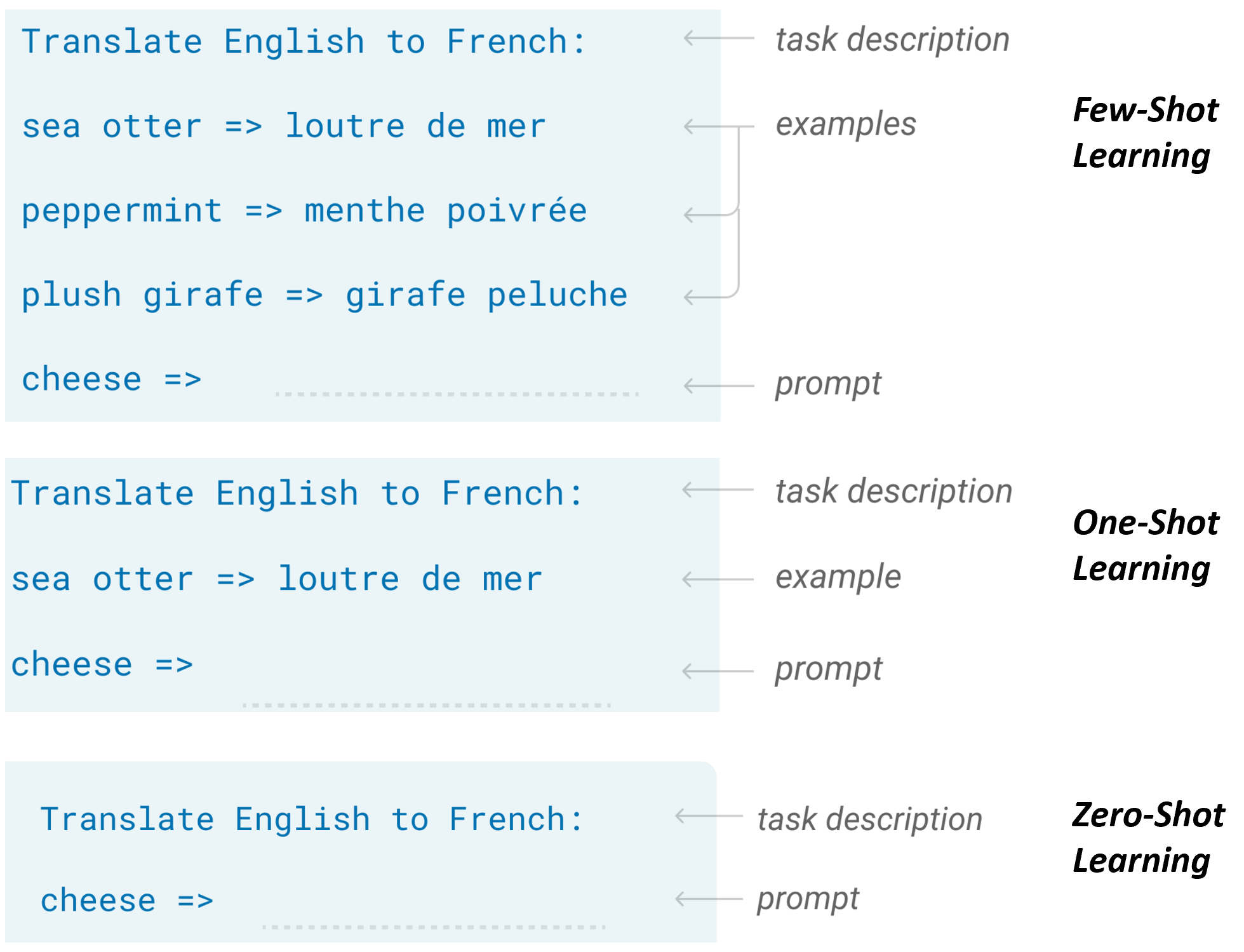}
    \caption{The emergent phenomenon of \emph{in-context learning}: It enables solving tasks the model was not explicitly trained for using a textual description and possibly examples (``shots''). \footnotesize{Adjusted from \ci{bro20}}}
    \label{fig:incon}
\end{figure}

Foundation models can serve as the foundation for many downstream applications, and they can be adjusted (or fine-tuned) with limited labeled training data for a specific task. Surprisingly, they can also be used to address tasks not explicitly trained for. They permit ``in-context learning'', i.e., during the training phase, the model extracts a rich set of patterns and broad skills from the diverse training data. In turn, a (language) model can perform downstream tasks simply by providing a \emph{prompt}, i.e., a description of the task in natural language and, possibly, a few examples. For example, GPT-3 can translate text as illustrated in Figure \ref{fig:incon} or add numbers by prompting, e.g., ``4+5 = '', although it was not trained on any mathematics task. Furthermore, performance can be improved by crafting adequate prompts. For example, \cite{koj22} showed that by simply adding ``Let's think step by step.'' to an input, the model performed significantly better on various benchmarks. This \emph{emergent behavior} is an unexpected phenomenon even for experts. It has not been observed in smaller models trained with fewer data. Still, foundation models are not yet outperforming existing models on all tasks or can be said to come close to artificial general intelligence (AGI) or human reasoning -- despite media reports on attributing them properties such as conciousness\footnote{\url{https://www.nytimes.com/2022/07/23/technology/google-engineer-artificial-intelligence.html}}. For example, the aforementioned mathematical skills are mostly limited to two digits addition and subtraction. Already for three digits or other operations such as multiplication, accuracy drops below 80\%.


\begin{figure}[htb]
    \centering
    \includegraphics[width=0.9\textwidth]{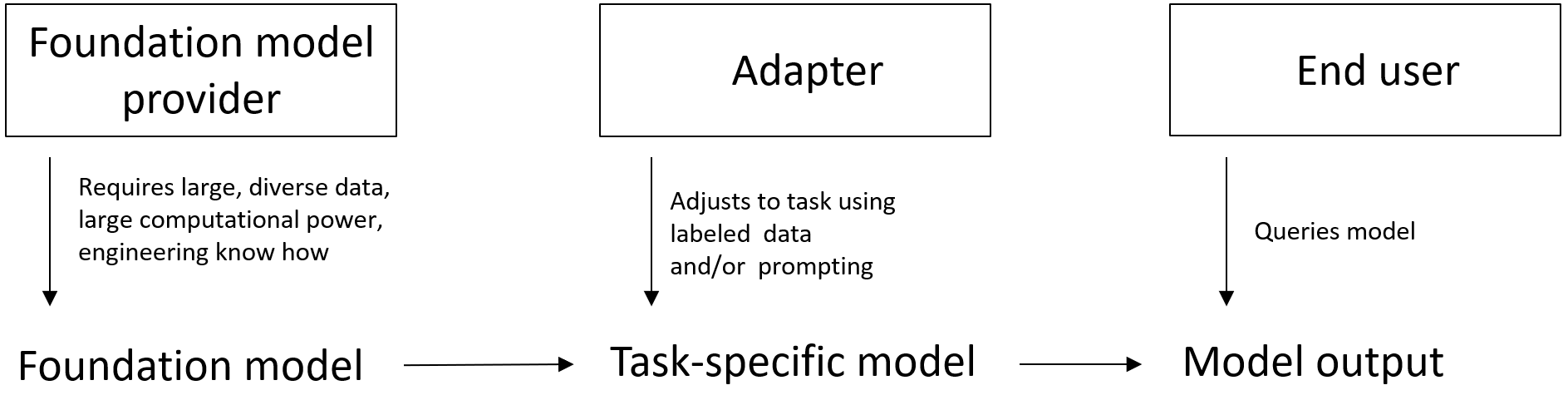}
    \caption{Actors and outcomes in AI development involving foundation models.}
    \label{fig:actor}
\end{figure}

Still, compared to prior deep learning models, foundation models not only strike through their breadth of applications but often through their performance. They commonly achieve state-of-the-art results on specific tasks outperforming tailored models. Furthermore, they tend to require less task-specific (labeled) data, which makes them attractive since labeled data can be costly. Their success is not because of technical improvements but rather due to the sheer size of models and the diversity and amount of training data. They also foster a paradigm shift from learning from a labeled dataset (supervised learning) to learning from a larger, unlabeled dataset in a self-supervised manner. \emph{Self-supervised learning} learns from task-agnostic data with self-generated labels (rather than from human annotators). Labels originate from artificially generated tasks. For example, BERT is trained to fill in missing words in a sentence (and two predict whether two sentences are in the correct order or not). During trainings a few words of a sentence are randomly removed and function as labels to be predicted.  Surprisingly, the resulting model is not just useful for predicting missing words but forms the basis for many other NLP tasks such as question answering, sentiment analysis, abstract summarization, etc.. Thus, foundation models can also be seen as models that transform data into an universal representation useful for many tasks as proclaimed, e.g., by IBM's Vice President for research\footnote{\url{$https://www.youtube.com/watch?v=_XR08GA9YH0$}}. 



Foundation models might transcend all areas of AI through \emph{homogenization} and lead to a concentration of power. Homogenization can occur in three ways, new AI models are adjustments of (i) a few foundational models (ii) trained on a few datasets (iii) by a few organizations. Foundation models might reduce deep learning model diversity since there is less need to encode task-specific information in a model by model developers, but general models can be employed due to the increased richness of training data. Foundation models might be trained on all globally available data, e.g., the entire Internet or large proprietary datasets such as images on social media platforms used by billions of people such as Instagram. This is costly, technically demanding, and requires access to large data, implying that only a few organizations are capable of doing so. For example, the training of GPT-3 is estimated to cost millions of dollars\footnote{\url{https://lambdalabs.com/blog/demystifying-gpt-3}}. However, these models can be adapted by little (labeled) data to specific tasks. There, the actors in AI development might change. AI developers could be partitioned into foundation model providers and their adapters as illustrated in Figure \ref{fig:actor}. Providers and adapters could be different entities. Furthermore, foundation models might also perform a variety of tasks without adaptation of any parameters.


Only in late 2021, a group of more than 100 associates of the human-centered AI institute at Stanford introduced and defined the term \emph{foundation model} as: ``A foundation model is any model that is trained on broad data (generally using self-supervision at scale) that can be adapted (e.g., fine-tuned) to a wide range of downstream tasks''\ci{bom21}. The definition was part of an extensive report of more than 200 pages, reviewing literature and elaborating on opportunities and risks of foundation models covering also technological and societal impacts and issues.
This paper distills the report\ci{bom21} and adds new perspectives (and literature). We more explicitly discuss the impact on organizations and human-AI interaction. Furthermore, we portray foundation models as a historical evolution of machine learning and articulate its difference in a compact and concise way. 

\subsection*{History of machine learning: From expert systems to foundation models} \label{sec:evo}
\begin{figure}[htb]
    \centering
    \includegraphics[width=\textwidth]{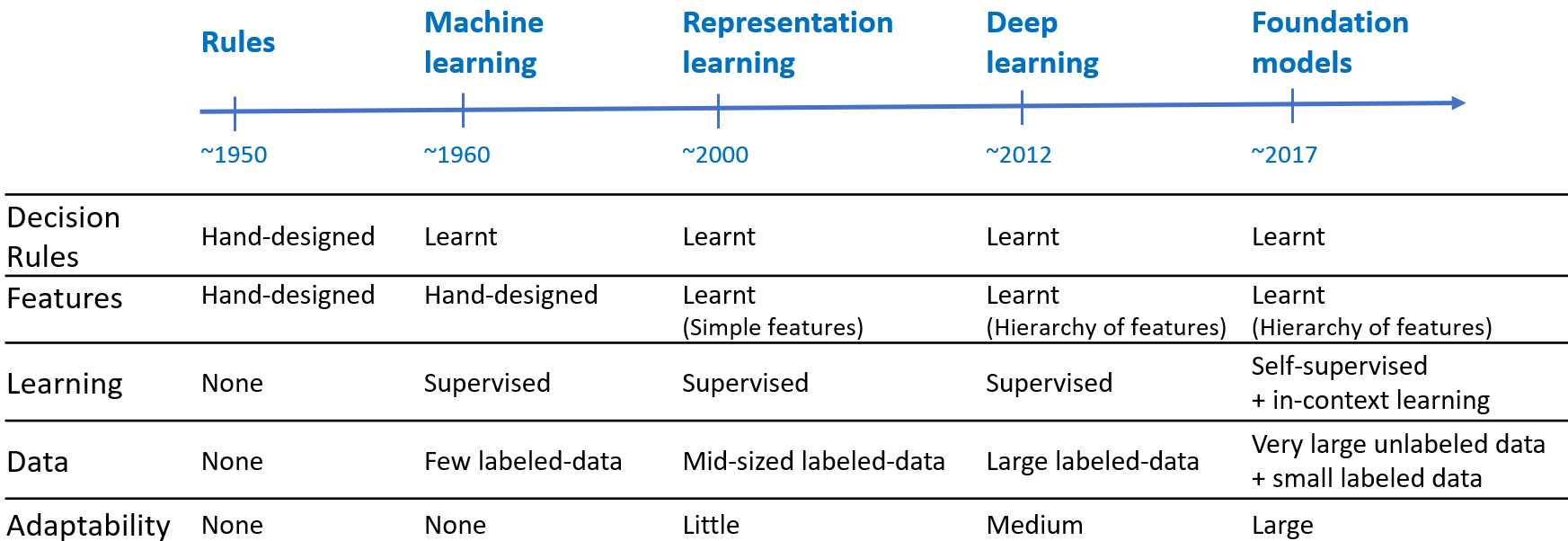}
    \caption{History of machine learning}
    \label{fig:hist}
\end{figure}

Foundation models can be seen as a ``logical'' step in the evolution of machine learning towards self-supervised, larger deep learning models as shown in Figure \ref{fig:hist}.

Before learning from data, decision-making by machines was done by expert systems\ci{jac86} starting with the introduction of computers in the 40ies. They were popularized in the 60ies through systems such as Dendral \ci{fei77} and also resemble early AI systems. The need for a detailed specification of how an input is transformed into an output made them costly and brittle. The next evolutionary step, i.e., machine learning, is associated with the term ``without explicitly being programmed'' coined by Samuel Jackson in 1959. Early machine learning learned decision rules based on given criteria relevant to decision-making, i.e., features. The definition of these features was subject to humans. So-called feature engineering requires domain and technical expertise (e.g., see \ci{low99,bay06} for examples in computer vision).  Representation learning aimed at automating feature engineering but was originally limited to relatively simple features. \\
\emph{Deep learning } achieved a breakthrough in representation learning. It reduced the need for feature engineering and allowed learning a hierarchy of features \ci[p.~1]{goo16}. Deep learning can be seen as an evolution of early neural networks starting in the 40ies\ci{mcc43,schm15,wang17}. Its success can be attributed primarily to more computation, more data, and an accumulation of technical innovations. It also allowed building models in a modular, flexible way by stacking various layers on top of each other. This makes it easy to enlarge models or combine models for data of different modalities. Deep learning transcended diverse areas of AI, including computer vision, speech, and natural language processing. Still, specific deep learning models were developed through different compositions of basic elements for a wide variety of tasks. The most prevalent paradigm for training models was supervised learning, i.e., training on a labeled dataset: Each input is associated with a label commonly provided by a human. Although models were developed based on specific datasets, learned features stored within a model can be reused across similar datasets through transfer learning (also called ``fine-tuning'')\ci{thr98}. For example, a model for computer vision such as Alexnet \ci{kri17} trained on the well-known ImageNet-1K\ci{den09} dataset with 1000 fixed categories could be adjusted to detect novel categories by fine-tuning the network using additional data containing samples of the novel categories. Fine-tuning often leads to better outcomes than training a network from scratch. Most real-world detection tasks share some commonalities, allowing to reuse parts of a deep learning model for other tasks. In NLP,  deep learning was used to estimate word-vectors\ci{mik13}. They provided a more flexible representation that could be used for a variety of NLP tasks. Still, researchers commonly adjusted models to almost any NLP task. This somewhat changed with the introduction of transformers \ci{vas17}. \\
\emph{Foundation models:} Transformers\ci{vas17} meant a breakthrough in NLP and constituted the birth of foundation models. Prior representations in NLP, such as word vectors, were ``static'' and context-independent, i.e., one-word vector per word, though most words' meaning depends more or less on context. For example, homonyms, i.e., words such as ``bank'' (financial institution, river bank, object to sit on), cannot be interpreted without context. Transformers allowed for a much more dynamic representation, i.e., contextual word embeddings. These embeddings take into account context, e.g., the sentence in which a word occurs.  Technically, transformers are built on existing design elements in deep learning, e.g., attention. A main success story of transformers, i.e., foundation models, in NLP has been text generation. For example, short texts generated from GPT-3 might be hard to be distinguished from those of humans\ci{cla21}. Before 2018, generating general-purpose language was seen as very difficult only approachable through other linguistic subtasks \ci{par13}. GPT-3 performs well on various tasks beyond text generation but shows mediocre results on others. For example, on the Naturalquestions benchmark, where the model should generate replies to questions posed on Google, such as ``Who played Tess on `Touched by an Angel'?'' it reached a score of about 30, while more specialized retrieval-based methods managed to achieve a score above 40 \ci{lew20,iza22}. However, the outperforming models\ci{lew20,iza22} themselves rely on variants of foundation models such as BART (a successor of the early foundation model ``BERT'').

A few years later, after the success in NLP, transformer variants\ci{dos20} also improved state-of-the-art results in computer vision. However, the gains were less profound. Leading models, i.e., convolutional neural networks (CNN), are better at learning suitable representations (relative to static word vectors in NLP). For image classification and object detection, self-supervised techniques \ci{hen21,chen20} are competitive to fully-supervised approaches. For example, CLIP\ci{rad21} matches the performance of systems trained on a labeled dataset, i.e., ImageNet-1k\ci{den09}, without using samples from the data for training. CLIP is trained on 400 million annotated images from the Internet, while ImageNet-1k consists only of about 1 million samples from 1000 categories.
Transformers have also entered other areas. Speech recognition systems are  increasingly built based on a large dataset of audio alone and then adapted with audio and associated transcriptions \ci{bae20}. Reinforcement learning\ci{jan21} has also seen the adoption of such models. They have also successfully leveraged multi-modal data, e.g., images and text. For example, systems for image generation based on textual descriptions such as Dall-E 2\ci{ram22} in conjunction with CLIP \ci{rad21} have shown stunning results. Here, Dall-E serves as a generator, while CLIP ranks the generated images based on their match with the textual description.  

Table \ref{tab:diff} provides a summary of the differences between foundation and non-foundation models.

\begin{table}[htb]
    \centering
    
    \begin{tabular}{c |c |c |c} \hline
      \textbf{Characteristic}  &\textbf{Characteristic}  &  \textbf{Foundation Model} &  \textbf{(Early) Deep Learning}  \\ \hline
       Model & Type  transformer &  CNN, LSTM, GAN \\ \cline{2-4}
             & Size & very large & medium-large \\  \cline{2-4}  
             & Adaptability & many, diverse tasks & few, similar tasks \\  \cline{2-4}
             & Prompting & Yes & No \\  \cline{2-4}
      & Training &self-supervised & supervised \\    \hline  
       Performance on& ... adapted tasks & State-of-the-art(SOTA) & high to SOTA \\ \cline{2-4}
             &  ... untrained task & medium to high & low \\ \hline             
       Data & Amount & very large & medium to large \\ \cline{2-4}
            & Labeled?& No & Yes \\ \hline       
    \end{tabular}
    \caption{Key differences between foundation models and conventional, smaller sale deep learning models}
    \label{tab:diff}
\end{table}


\section*{Applications}
Foundation models extend deep learning's applications and can enter almost any domain. Here, we only mention examples from a few areas: healthcare, biomedicine, law, and education.  A prominent area is \emph{healthcare}. As foundation models are capable of serving as an integrated knowledge reservoir, they can be queried and adapted to various individual tasks in healthcare. They can serve as an interface for healthcare providers, e.g., to retrieve relevant cases, suggest diagnosis and treatments \ci{ras21}, and as an interface for patients, e.g., answering patient questions related to preventive care \ci{dem20}.
In \emph{biomedicine}, drug discovery through designing molecules could be facilitated using the generative capabilities of foundation models \ci{Kad17}. There are also opportunities in personalized medicine by integrating multimodal patient data, including textual diagnosis onto medical images\ci{ouy20} as well as in improving clinical trials, e.g., by identifying eligible patients given multimodal patient data \ci{har19}. 
In \emph{law}, foundation models can help in providing legal information for citizens \ci{que20} or supporting lawyers in identifying relevant documents, e.g., patents or reviewing contracts \ci{hend21}. Criminal law risk scoring in parole decisions \ci{ber21}. In education, foundation models can provide feedback for students and assist teachers, e.g., through automatic grading \ci{sch22}. 

\section*{Technology}
Foundation models are deep learning models, i.e., currently large transformer models\ci{vas17}. They distinguish themselves less by technical innovation than by the sheer size of data and models. However, some technical design decisions and the fact that data is unlabelled play a role. For example, the training regime for foundation models is self-supervised training. 

\subsection*{Model: Transformer, attention and key-value pairs}
Foundation models mostly differ from deep learning models based on their expressivity (ability to capture diverse information) and scalability (ability to leverage large amounts of data both in terms of computational efficiency as well as being able to keep learning with growing amounts and diversity of data). Other relevant aspects that exhibit fewer differences to prior deep learning models include multi-modality (the processing of multimodal data such as images and text in one model), memory (representing and retrieving knowledge, possible from a model external source) and compositionality (modularity of the model)\footnote{This is a deviation from \ci{bom21} that regards compositionality as ``ability to generalize through its representation and model modularity''. We view compositionality more narrowly as ``model modularity''. }. Multimodality can be seen as part of expressivity, i.e., expressivity can be measured in terms of capturing information of one modality (e.g., very diverse images) and various modalities (e.g., ability to capture text, images, and audio). The rest of the section covers fundamental technical ideas of foundation models that are relevant for a deeper understanding of these models, but less technically inclined readers might proceed with the next subsection.\\
\emph{Expressivity }  describes the space of input and output relationships learnable by a model. As such, it relates to model complexity (more complex models can typically capture more diverse relationships) and inductive bias, e.g., encoding of prior assumptions into the model. Inductive bias is often based on the structure of the data and how data should be processed to obtain outputs. Generally, the less data is available for a task, the more inductive bias is preferable, i.e., the more task or data-specific knowledge should be encoded into the model. Foundation models typically require large expressivity to leverage large and diverse training data. They need little inductive bias since all information should be abundant in the data. Inductive bias might even be harmful. For example, for CNNs built on the idea of locality, i.e., to detect an object, the model considers small patterns in a small patch (often 3x3 pixels) and grows  the area with each layer which might constitute a pattern or object, i.e., the receptive field. This approach assumes that long-distance relationships of image parts are less relevant. This assumption seems mostly true, as confirmed by the excellent performance of CNNs, i.e., most objects pose a single, localized unit. But it has been put into question due to \emph{transformers}, the leading architecture for foundation models. Transformers for computer vision applications such as Dall-E do not process images by processing small patches as CNNs, but rather look at a few large patches (e.g., 16x16 pixels rather than 3x3). Outputs for these large patches are then combined to derive outputs. Transformers might be seen as fully connected networks leveraging a few well-known mechanisms of deep learning, such as attention, residual connections, and normalization. \emph{Attention} is a weighing or gating scheme of inputs. Attention seeks to produce outputs that are either close to 0 or 1. That is, they form a gate that can be closed (0), not letting any information through or open (1) allowing information to flow or ``somewhat open''. The computed attention weight is multiplied by an input so that the input can flow (if attention is 1) or is stopped (if attention is 0). Attention allows, for example, to easily express behavior: ``if the background is blurry and the center of an image is sharp, the most relevant pixels to recognize an object must be in the center''. In this case, attention could set weights for blurry input areas to zero and others to 1. Thus, attention can serve as a gate or mask that eliminates unimportant inputs from further processing in upper layers that could otherwise mislead the model. Another important architectural aspect is learning key-value pairs rather than learning weights that resemble patterns matched with the input. That is, in CNNs (and fully connected networks), an input is compared to a pattern (encoded in learned) weights by applying a dot product. The output is the similarity or match between the two. For key-value pairs, each key is associated with its own value. Similarities between all keys and inputs (often called queries) are computed. The similarity of a key and an input determines the contribution of the key's associated value to the output. If a key is an exact match with an input and all others keys match poorly, its value will be returned. If all keys match equally well, the returned value is the average of all values.
\emph{Multi-modality } might be seen as expressivity of a model with respect to multiple modes of data. In a real-world scenario, information can be captured in multiple modes, e.g., visual, audio, or textual. Using all modes together commonly leads to better outcomes. Design dimensions for foundation models are (i) multimodal interaction (degree of weight sharing when processing modes) and (ii) generality and specialization for each mode. One might extract high-level features of audio (such as words spoken) and video (such as objects shown) before linking the (processed) data of each mode. In this case, models can be specialized towards each mode, which allows to leverage mode-specific data properties through their inductive bias. However, linking the models at lower levels can be seen as a more general approach that ``lets the data speak for itself'' and lead to better outcomes.
\emph{Memory: } A model can contain all knowledge in its parameters or (also) resort to a separate memory, e.g., to store facts. Separate memory might be more interpretable, easier to update and reduce model size. For example, retrieval-based question-answering models look up relevant documents for further processing based on transforming input queries using foundation models. They have outperformed non-retrieval-based models while being an order of magnitude smaller \ci{iza22}. 

\emph{Scalability } expresses the ability of a model to be easily adjustable to larger amounts of data. This encompasses two characteristics. First, the ability to benefit from larger amounts of data. That is, more data lead to better model performance. For deep learning, this can often be achieved by enhancing model size in terms of width and depth, e.g., adding more layers (of the same type) and increasing parameters per layer. However, not all models are ensured to be easily scalable in this manner, e.g., models might become difficult to train if they become very deep (vanishing gradient). Furthermore, models should also (remain) easy to adapt, e.g., ideally, a model trained on a larger amount of broad data should require less task-specific information to fine-tune to a specific task. The second concern of scalability is the ability to process more data in a reasonable time. Generally, when increasing data, one aims for a roughly linear increase in time, i.e., doubling the training data should also only double the training time. A linear increase is not guaranteed since more data also requires larger models, i.e., more processing per sample, and large models might not fit on a single GPU, thus, adding overhead due to distributed training. For large foundational models, a key question is whether a model can effectively be trained in a parallel manner (e.g., using multiple GPUs). For example, sequence models are ineffective for large inputs, i.e., very long sequences, since they process one element of the sequence after another without any opportunity to parallelize. In contrast, for a CNN, each layer can be parallelized easily.

\emph{Compositionality } touches upon architectural properties and modularity of models and representations. Deep learning models are highly modular, i.e., a layer can be added or removed in an easy manner, often without a very large performance impact. Modularity has implications on other aspects, such as interpretability (systems with well-separated components and few interactions are usually easier to interpret). While interactions among layers are generally large, making interpretability difficult, they can be low in some cases. For example, graphs learn nodes and relationships between them. 

\subsection*{Training and Adaptation}
Foundational models are first trained with a large dataset before possibly being adapted with a smaller dataset. Training of a model should leverage the broad data and ensure that the learnt model captures information that is (domain) complete, i.e., includes all information needed to capture a wide variety of tasks. A model trained on text should be adjustable to a wide variety of NLP tasks. Furthermore, it should be scalable, i.e., remain efficient with growing amounts of data and larger model sizes. \\
Even in case foundation models can solve a task without further adjustments, additional adaptation often improves performance. The choice of adaptation mechanism depends on three factors: compute budget, data availability, and access to foundation model. Adjusting all parameters of a foundation model is only feasible if one has access to model internals, i.e., its parameters and gradients. In a cloud computing setting, a foundation model provider might only provide model outputs. Adjusting all of the billions of parameters also requires a sufficient computing budget, access to model gradients, and a sufficient amount of data. Adaptation can be based on transfer learning, i.e., adjusting or adding layers. In-context learning allows using a foundation model without any further additions or changes to the model for novel tasks. It can also improve model performance through prompting, e.g., designing inputs or altering. Although research on prompting is growing at a rapid pace\ci{liu21}, it is not yet understood well.  

\subsection*{Evaluation}
Evaluation of foundation models is difficult since they are not designed for a specific task with a well-defined metric. Performing an intrinsic, task-agnostic evaluation with appropriate measures is hard since it is often unclear what the model's applications are when it is evaluated. An external evaluation based on performance on more narrow, downstream tasks is more tangible but also has shortcomings. For example, for a language model aiming to predict the probability of a word in a sequence, one might use a measure such as perplexity. Extrinsic evaluation entangles the evaluation of the model and the adaptation mechanism of the foundation model and the task. That is, each foundation model might require a different adaptation for a specific task. In turn, differences in extrinsic evaluation might be due to the adaptation procedure rather than due to the model.  
Evaluation covers various theoretical constructs such as accuracy, robustness, fairness, data, and computational efficiency for training and adaptation). Evaluation should be undertaken separately for the foundational model and the adapted model. A foundational model might be trained once and then adapted to many different tasks. Thus, computational costs for training and testing the foundational model and for the adaptation of the model should be reported separately. Similarly, measures for data efficiency should be reported separately for training of the foundational model, where data is multiple orders of magnitude larger than for adaptation. For adaptation, in particular, for prompting zero, one-shot, and few-shot learning play a relevant role as well as diversity. Thus, there might be a shift in evaluation for foundation models compared to earlier deep learning, i.e., from a single, large validation set to evaluating on many tasks with few samples.
Furthermore, standardized datasets should be used to ease model comparison. Due to the size of the training dataset and the costs to obtain such a dataset paired with commercial interest (research labs in industry are currently leading the development of foundation models) this remains a problem. This can lead to unfair comparisons and wrong conclusions since a single measure is insufficient to judge a model. For illustration, a very good model might be outperformed in accuracy by a poor, highly inefficient model trained with much more data and for a longer time.

\subsection*{Security, Privacy, and Robustness}
Foundation models posit multiple risks. They constitute a single point of failure. That is, if a model has memorized private data of an organization or an individual, all adjusted models could potentially leak this sensitive information. Data poisoning, i.e., the intentional introduction of data samples into training data to manipulate model behavior is a key concern since data might be scrapped from public sources such as the Internet. The generality of foundation models also makes them more susceptible to unintended, unforeseen use (called function creep or dual use). For example, CLIP was trained on image-text pairs. While its model card puts surveillance as out-of-scope, the model has also learnt to capture rich facial features\ci{goh21} and could be used for this purpose.
Foundation models also offer opportunities to improve security and robustness. Large model size and large training data tend to be positively associated with robustness \ci{bub21,sch18} to distribution shifts and adversarial samples. Foundation models tend to be more robust to data drift, e.g., for specific benchmarks\ci{rad21} and real-world data \ci{kum22}. Adapted models might also inherit (increased) robustness from foundation models. 


\subsection*{AI safety and alignment}
For foundation models AI safety concerns are similar to ordinary deep learning, but more profound due to their more widespread adoption and their unforeseen applications. This increases the risk of misalignment between the training objective and the wanted behavior. For example, a common objective for general-purpose language models is to predict the next word within diverse documents. But the desired goal can be to output only true or helpful text\ci{tam21}. Furthermore, failures of AI, e.g., due to a lack of robustness, can be more severe for foundation models due to their larger diffusion. 

\subsection*{Interactive systems}
Due to being trained on large amounts of data and their generative capability, deep learning and, more so, foundation models can form the basis of interactive systems such as chatbots\ci{ada20}. In addition, foundation models can foster interactivity due to their ability to ``in-context learning'' encouraging to alter or design prompts, i.e., a user can query a model, investigate the output, adjust the query by providing an example, etc. However, prompt engineering is not yet understood and is subject to research\ci{liu21}. Furthermore, it is not clear how to best design interactive systems involving prompting. Even very basic questions such as ``Should one attempt to use XAI to foster a deeper model understanding or just display outputs for a query?'' have not been studied. Prompting blurs the line between users and developers, as users can instruct how the system should behave through instructions and examples. This is a new interaction pattern to which users must learn to adhere. In summary, key differences for systems leveraging prompting compared to typical commercial conversational agents such as Amazon's Alexa or Siri are the provisioning of examples within a query and that, commonly, the interaction is about creating an artifact, e.g., a generative system like Dall-E should come up with illustrations based on user queries.

\subsection*{Other areas}
Foundation models impact many other areas, such as systems, data, and XAI, where they give rise to new questions. However, there is arguably a larger overlap with existing issues for classical deep learning.
Foundation models come with high computational and memory demands, which in turn, impacts computer \emph{systems}. For example, they might lead to co-design of models, development software, and hardware, e.g., to foster parallelism and optimize specific mathematical operations. Due to the resource costs in developing these models, there is also pressure to automate tasks during development. In productionization foundation models, additional factors such as overall system latency (e.g., how long does it take to process a query?) and automated monitoring and testing of a system (e.g., how to ensure that model performance does not degenerate?) also play an important role.
Foundation models put an increasing emphasis on \emph{data}. In turn, this increases and poses new challenges to data management at scale (especially data quality), integrating data across modalities and governance regulations.
\emph{XAI} becomes more demanding as models grow in size and data becomes more versatile and often multimodal.

\section*{Societal Impact and issues}
Complications in understanding societal impact arise due to the intermediary nature of foundation models, making accountability more difficult, e.g., issues can stem from training and adaptation data, the original model and adapted model, etc.
\emph{Inequity, fairness, and harms}: The fact that foundation models form the basis of many other models leads to two types of biases: Intrinsic biases that indirectly but pervasively affect downstream applications and extrinsic harms that arise in a specific downstream application. For example, representational bias can be an intrinsic bias of a model, e.g., African Americans might be underrepresented, while the most prevalent language is likely to dominate. Training with unbalanced data while dealing with language variation is a challenge\ci{ore19}. This intrinsic bias can lead to extrinsic harm, e.g., a performance disparity on text in African American English might negatively affect black people\ci{koe20}. However, when the task is to describe plants, the intrinsic bias might not lead to extrinsic harm. Sources of biases are data, models, and modelers. 
Mitigating inequities aims at interventing at different steps of the model development pipeline (e.g., data \ci{luk20}, model objectives \ci{zha18}, and adaptation). It is challenging since the possible uses of foundational models are unknown, and issues might stem from various sources, e.g., from data used to train as well as to adapt a foundation model. In turn, this makes it more difficult to assign responsibilities, and it makes a proactive intervention difficult. Recourse, i.e., mechanisms for quickly adjusting due to feedback from end-users or entities adapting models, might be necessary.
Furthermore, intervention and recourse deserve attention. Due to these models' widespread and unforeseeable use, it might be difficult to identify all issues before deployment. Challenges include how and where to intervene, e.g., identifying sources for bias or harm and how to recourse. That is, in case of harm or inequity, a framework for resolving a recourse for the harmed parties is missing.

\emph{Misuse and improper usage}: Misuse can arise due to increased content quality, leading to content that is often indistinguishable from those of humans\ci{cla21} at low costs reducing the barrier for harmful attacks\ci{bru18}. Additionally, personalization of content is easier. It is difficult to anticipate all possible uses of foundation models, and as such unforeseen, improper use and misuse are possible.  Furthermore, GPT-3 itself, if not used carefully, can contribute to misinformation. For example, given a title (and a subtitle) GPT-3 generated newspaper articles that humans could not distinguish from human-written articles, i.e., they reached merely an accuracy of 52\%, while guessing already gives 50\%\ci{cla21}. Thus, generation of high-quality texts becomes easy. An article from the original GPT-3 paper\ci{bro20} has been analyzed for the truthfulness of facts. The generated article on ``United Methodists'' contains a number of incorrect facts, e.g., the article claimed that the church split in 1968 though actually, a merger occurred. It also claimed, among other wrong claims, that the church's general conference occurs annually, though it takes place every four years. 

\ifbis
\else
Created content for misuse can refer to artifacts such as images or text as well as to explanations for decisions that could be deceptive\ci{sch20d}.  At the same time, foundation models can also serve as detector for harmful content originating from humans as machines.
\fi


\emph{Environment}:
But due to their large model size and data size, foundation models cause high energy consumption during training and operation. Models like GPT-3 or Gopher produce more than $CO_2eq$ than a roundtrip flight with 300 passengers from New York to London, i.e., more than 300 $tCO_2eq$ \ci{rae21}. Another aspect of foundation models is their energy demands due to their computational and data needs. In turn, this raises issues with respect to sustainability, in particular, energy consumption. However, as foundation models are used for many applications, training and development costs might be amortized. That is, it might be more efficient to train a large foundation model and adapt with few data and computation to many applications than to develop many mid-sized, specialized models. Also, some techniques to reduce energy consumption used in classical deep learning\ci{sch22b} such as reducing hyperparameter optimization energy costs by using smaller models trained on a subset of the training data and then extrapolating performance is more challenging since some model behaviors only ``emerge'' when training on large amounts of data. 

\emph{Economics and Organizational Impact}
Foundation models can be thought of as what economists refer to as a general-purpose technology\ci{bre95}. They might increase productivity and innovation and boost the adoption of AI. They might reduce the need for labeled data, which is an important cost factor. The homogenization (and standardization) of training procedures, combined with cost-per-use models and cloud deployments, makes it easier even for small companies with a limited financial budget, know-how, and data to leverage such models. 

They might also enable new applications and disrupt others. For example, the capability of these models to generate high-quality content at low costs offers the possibility to create illustrations, where they are not yet commonly used and to alter the process of how they are generated. For instance, DALL-E \ci{ram22} could change the market for illustrations, similar to how inexpensive cameras revolutionized photography. Models like DALL-E allow a user to specify their design in natural language and receive proposals in seconds, adjust their description depending on the proposal, and repeat. This requires an end-user to be skilled in effectively using the system. In contrast, such a process is much slower and more expensive with human designers.

Centralization of data and the resulting foundation models can lead to fewer people having economic mobility and opportunity \ci{bry22}. Such a market concentration implies a shift of power and decision rights. Open science efforts aim to make such models more accessible through a joint community effort. Still, this organizational structure raises new challenges, e.g., how to perform efficient distributional training (low latency and high bandwidth needs make training models residing within a network of contributors challenging). Proprietary access to data of companies poses another challenge.
While deep learning vastly benefited from a culture of openness, where models and data were made publicly available, foundation model might reverse this trend. Data (e.g., for GPT-2) has not been released, and models are either not released at all or only accessible through an API (e.g., GPT-3) for a limited number of people. Furthermore, even if models and data were released, many researchers would not have access to the computational infrastructure or possess the engineering skills to handle them. The research community is attempting to train such foundation model like GPT-3 through efforts such as EleutherAI\footnote{\url{https://www.eleuther.ai/}}. But overall, the disparity between industry and academia is concerning, especially given the focus of a company on profit rather than society as a whole.

On an organizational level, for a company, dependencies on customers and suppliers of AI-related technology might change. That is, a company might rely on a foundation model provider with proprietary models (see Figure \ref{fig:actor}) rather than training models from scratch or using publicly pre-trained models. With increased adaptation to a foundation model dependencies can increase, while at the same time, the fact that adaptation tends to require fewer data and training effort might also make it easier to switch between foundation model providers. Governance of models becomes more intricate. It might be more difficult to assess model risk and behavior. For once, model access might be limited, e.g., if only outputs are accessible, certain explainability methods that require internal model information, such as gradients, cannot be used. Data access might also be limited, making it difficult to assess data quality and data biases. Power and knowledge asymmetries can also contribute to risks. Since foundation model providers are usually large companies with a lot of expertise, smaller companies with less knowledge of models and fewer resources might find it hard to prove that they are not accountable for certain model behaviors. 

\emph{Ethics of scale and Legality}
Homogenization can amplify bias and arbitrary exclusion \ci{cre22}. It can also lead to ``culture homogenizing'', i.e., spreading one implicit perspective across multiple domains. The incentive for mass data collection and foregoing privacy rights grows. Norms for foundation model development are also important and need to be fostered, e.g., to ensure that developers consider issues and implications arising from foundation models. Foundation models might also lead to a shift in auditing and release. Access to foundation models is more limited, e.g., due to a lack of technical and computational capabilities (in case models are released) or due to proprietary datasets and models kept secret by companies. \\
The use of large, uncurated data increases the risk of privacy and copyright issues. For outputs, liability issues can arise as well as legal protections for outputs touching on ownership and questions like if "AI speech" is covered under the First Amendment protecting freedom of speech. For foundation models the relevance of these questions is bigger and more intricate since the content is of higher quality and they are more widely adopted. Also, there are more actors than in a classical ML setting, i.e., the foundational model provider and the adapter.

\section*{Conclusions and Future Research}
Foundation models can be seen as a major step forward in AI through scaling up data and models (similarly to how deep learning meant a breakthrough for AI). Many opportunities and issues overlap with deep learning, e.g., challenges in XAI\ci{mes22}, AI in business \ci{ngu22}, and ethical concerns such as fairness\ci{feu20}. However, for foundation models, answers to these questions might differ and, additionally, many new questions emerge by considering key aspects of foundation models. Their foundational character stems from (i) performance boosts on certain tasks, (ii) their adaptability to many tasks, (iii) homogenization (``access, power and control of these models''), (iv) novel emergent behavior (``in-context learning''), and (v) resource needs. Each of these characteristics can be investigated from various perspectives, such as organization, technology, people/society, which, in turn, gives rise to a diverse set of research questions.\\ 
Gains in performance and their adaptability to diverse tasks with few labeled data might foster the adoption of AI and enable new applications. For example, systems like Dall-E generate illustrations of unpreceded quality based on textual descriptions. Exemplary questions are: ``What are business models and services emerging from new applications enabled by foundation models? How is intellectual property handled and attributed?'' Here, existing works, e.g., on AI as a service \ci{lin21}, might provide a starting point.

Homogenization leads to new power and control structures. AI might shift from a technology characterized by openness, where pre-trained models are shared even by research institutions of industrial companies, to a technology that becomes inaccessible to researchers and controlled by a few organizations. Furthermore, the actors in delivering AI services might change from two to three, e.g., AI products might rely on foundation models provided by one company that are further adapted by another company. Foundation models raise many organizational questions, e.g., ``How to govern data, foundation models, and adapted models? How to ensure accountability?'' Existing data and AI governance frameworks might be leveraged, e.g., \ci{sch22g}.\\
Novel emergent behavior, i.e., ``in-context learning'', allows for new interactional patterns with AI, such as prompting. Combined with the ability of foundation models to perform a diverse set of possibly unforeseen tasks, this raises questions such as ``How to prevent and detect misuse? What are design principles for systems where interaction between humans and a foundational model occurs?''. Guidelines for human-AI interaction could be further developed to this end \ci{ame19}.\\
Resource needs are entangled with homogenization and might be seen as a root cause. We believe that tackling resource needs might not only help in making these models accessible to a wider audience of researchers, businesses, and end users but is also a must to combat our climate crises. Thus, we define key questions as ``How to reduce data and computation needs? General design principles in green data mining and machine learning, such as reuse, reduce, and support \ci{sch22b}, could serve as a starting point for investigation. \\
More bluntly and forward-looking, one might ask ``What comes after foundation models?''. Our historical analysis hints at an obvious trend: ``Larger, more compute-intensive models trained on even more data of larger diversity''. Even larger models might be more versatile through adaptation and prompting, leading to broader adoption in practice. Thus, research should tackle pressing issues related to foundation models as soon as possible.

\ifbis
\else
\bibliographystyle{plain}
\fi

\bibliography{references}

\end{document}